\documentclass[10pt,twocolumn,letterpaper]{article}

\usepackage{cvpr}
\usepackage{times}
\usepackage{epsfig}
\usepackage{graphicx}
\usepackage{amsmath}
\usepackage{amssymb}
\usepackage{multirow}
\usepackage{booktabs}
\usepackage{flushend}

\usepackage{algorithm}
\usepackage{algorithmic}

\usepackage{amsmath}
\DeclareMathOperator*{\argmax}{arg\,max}

\usepackage[breaklinks=true,bookmarks=false]{hyperref}

\cvprfinalcopy 

\def\cvprPaperID{****} 

\begin{document}

\title{Activitynet 2019 Task 3: \\Exploring Contexts for Dense Captioning Events in Videos}

\author{Shizhe Chen\textsuperscript{1}, Yuqing Song\textsuperscript{1}, Yida Zhao\textsuperscript{1}, Qin Jin\textsuperscript{1\thanks{Corresponding author.}}, \\
Zhaoyang Zeng\textsuperscript{2}, Bei Liu\textsuperscript{2}, Jianlong Fu\textsuperscript{2}, and Alexander Hauptmann\textsuperscript{3}\\
\textsuperscript{1} Renmin University of China, \textsuperscript{2} Microsoft Research Asia,  \textsuperscript{3} Carnegie Mellon University\\
{\tt\footnotesize \{cszhe1, syuqing, zyiday, qjin\}@ruc.edu.cn, \{v-zhazen, bei.liu, jianf\}@microsoft.com, alex@cs.cmu.edu}
}

\maketitle

\begin{abstract}
Contextual reasoning is essential to understand events in long untrimmed videos.
In this work, we systematically explore different captioning models with various contexts for the dense-captioning events in video task, which aims to generate captions for different events in the untrimmed video.
We propose five types of contexts as well as two categories of event captioning models, and evaluate their contributions for event captioning from both accuracy and diversity aspects.
The proposed captioning models are plugged into our pipeline system for the dense video captioning challenge.
The overall system achieves the state-of-the-art performance on the dense-captioning events in video task with 9.91 METEOR score on the challenge testing set.
\end{abstract}

\section{Introduction}

The task of dense video captioning \cite{krishna2017dense} aims to generate a sequence of sentences to describe a series of events in the video.
A typical framework for dense video captioning \cite{chen2018ruc+} is based on two stages: 1) event proposal generation to detect potential events in the video, and 2) event caption generation to produce a sentence for each specific event.

Though dense-captioning events in the video is similar to the traditional video captioning task which generates a sentence for a video clip, directly deploying traditional video captioning models leads to poor performances due to lack of context in the video \cite{krishna2017dense}.
Being aware of contexts of an event not only can provide holistic information to understand the event more accurately, but also tell differences between target event and its context to generate more diverse captions. 
Therefore, previous endeavors have employed different contexts for the event captioning. 
Krishna \emph{et al.} \cite{krishna2017dense} propose to generate event proposal first and then dynamically select neighboring events as context for target event captioning.
Our previous work \cite{chen2018ruc+} has proposed to implicitly encode global video contexts into each segment feature via LSTM, as well as explicitly employ local temporal regions of the target event as contexts.
Besides visual contexts, Mun \emph{et al.} \cite{Mun_2019_CVPR} further consider sentence contexts from previous captions to improve diversity and coherency.
However, there are no comprehensive evaluations on contributions of different contexts for event caption generation.

In this work, we systematically explore and compare different contexts for dense-captioning events in video.
We design five types of context including segment-level contextual feature, local context, global context, event context and sentence context.
Two broad categories of event captioning models are proposed to employ different contextual information, namely intra-event captioning models and inter-events captioning models.
We carry out extensive experiments on the ActivityNet Captions dataset to evaluate the contributions of different contexts and captioning models from both accuracy and diversity aspects.
Our preliminary experiments suggest that inter-events models can generate more diverse event descriptions than intra-event models, but are slower and achieve slightly worse accuracy in terms of METEOR metric than intra-event models. 
Therefore, the intra-event models are more suitable for the evaluation of the challenge which focuses on the METEOR performance.
We plug the proposed intra-event models with contexts into our dense video captioning pipeline, and achieve state-of-the-art performance on the challenge testing set.

The paper is organized as follows. 
In Section~\ref{sec:context_models}, we introduce the proposed contextual types and event captioning models.
Then in Section~\ref{sec:densecap_system} we describe the overall pipeline of our dense video captioning system for the challenge.
Section~\ref{sec:expr_results} presents the experimental results and analysis.
Finally, we conclude the paper in Section~\ref{sec:conclusion}.

\section{Event Caption Generation with Contexts}
\label{sec:context_models}
The context has played an important role for dense video captioning.
Although previous works \cite{krishna2017dense,chen2018ruc+,Mun_2019_CVPR} have employed different contexts for event caption generation, no systematic evaluations on the contribution of different contextual information has been investigated.
In this section, we make a thorough comparison of different contexts and modeling approaches for dense event captioning.

We mainly divide the contexts for dense event captioning into five categories as follows:
\begin{itemize}
	\item Segment-level contextual feature: enhance segment-level feature with local or global video contexts such as $V_c$ features from LSTM described in Section~\ref{sec:segment_feature}. Such features are event agnostic, and contain larger temporal receptive fields than isolated segment-level features.
	
	\item Local context: encode temporally neighboring video contents for target event captioning. The local neighboring regions can provide necessary antecedents and consequences for understanding an event. Such context only relies on the target event itself to compute.
	
	\item Global context: encode the global video content except the target event as context. It provides an overall picture of the whole video outside the event. 
	
	\item Event context: encode neighboring events of the target event as context. It considers the correlations of different event proposals. The difference between event context and local context is that the local context does not necessarily to be potential events that capture complete actions. So local context does not require to know other events in the video while the event context does.
	
	\item Sentence context:  encode generated event sentence descriptions as context. It considers what has been said in the past events for the target event captioning, which aims to improve diversity and coherency of the generated event captions. So it requires to know the past events and past generated captions.
\end{itemize}

The first three types of context do not rely on other events except the target event, while the latter two contexts requires to be aware of other detected events in the video.
Therefore, in order to employ the above contexts, the event captioning models can be categorized into two types, namely intra-event models and inter-events models.

\begin{figure}
	\centering
	\includegraphics[width=\linewidth]{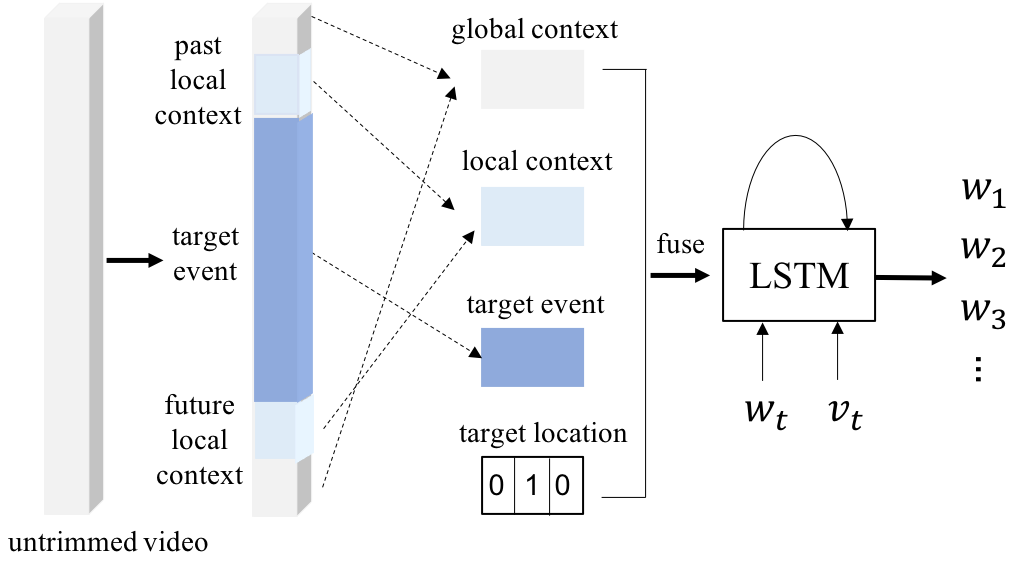}
	\caption{The architecture of intra-event model for event captioning with local and global contexts.}
	\label{fig:event_independent_models}
\end{figure}

\begin{figure}
	\centering
	\includegraphics[width=\linewidth]{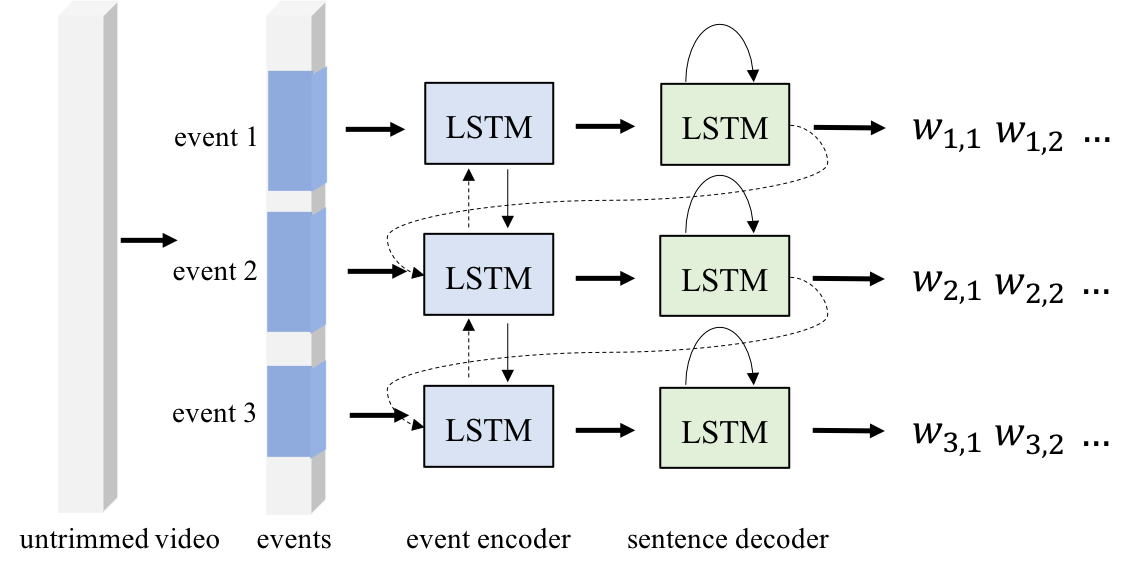}
	\caption{The architecture of inter-events model for event captioning with event and sentence contexts.}
	\label{fig:event_dependent_models}
\end{figure}

The intra-event model is illustrated in Figure~\ref{fig:event_independent_models}, which can employ either segment-level contextual features, local context and global context or their combinations.
It only requires the target event proposal for generation, so that different event proposals in one video can be processed in parallel for speed up.
The inter-events model however requires the existence of other events in video for target event captioning as shown in Figure~\ref{fig:event_dependent_models}.
We propose an event encoder to capture contexts from different events, which can be uni-directional in on-line setting or bi-directional in off-line setting.
Then a shared sentence decoder is employed to generate event descriptions for each event.
To capture sentence contexts, the generated caption from the previous event can be further fed into the event encoder, the structure of which is similar to Mun \emph{et al.} \cite{Mun_2019_CVPR}.
Therefore, the inter-events models with sentence contexts are processed in sequential order which cannot be paralleled in training and inference and leads to slower speed.
In this work, we utilize the temporal attention sentence decoder \cite{yao2015describing} which dynamically attends to relevant temporal segment of the target event for event captioning.

\section{Dense Video Captioning System}
\label{sec:densecap_system}
The framework of our dense video captioning system is based on our previous endeavor \cite{chen2018ruc+} in ActivityNet Challenge 2018, which consists of four components: 1) segment feature extraction; 2) event proposal generation; 2) event caption generation; and 4) proposal and caption re-ranking.

\subsection{Segment Feature Extraction}
\label{sec:segment_feature}
We divide each video into non-overlapping segments with 64 frames per segment, and extract four sets of features for each segment.
The four sets are: 1) basic feature set $V_b$ that captures global content of each isolated segment from different modalities; 2) object feature set $V_o$ that captures fine-grained object-level features of the segment; 3) semantic feature set $V_s$ that represents each segment as semantic concept distribution; and 4) context feature set $V_c$ that captures contextual information for each segment.
In the following, we describe each feature set in details.

The basic feature set is the same as extracted features in our previous work \cite{chen2018ruc+}, which includes: 1) Resnet200 \cite{he2016deep} from image modality pretrained on ImageNet dataset; 2) I3D \cite{carreira2017quo} from motion modality pretrained on Kinetics dataset; and 3) VGGish \cite{hershey2017cnn} from acoustic modality pretrained on Youtube8M dataset.
These three features are temporally aligned and are concatenated together as $v_b^t$ for the $t$-th segment.
Therefore, the video is converted into $V_b=\{v_b^1, \cdots, v_b^T\}$.

The object feature set aims to capture more detailed spatial information in the segment. 
We utilize Faster R-CNN \cite{anderson2018bottom} pretrained on the VisualGenome dataset to detect objects in the mid-frame of the segment.
We only keep object classes that overlap with frequent nouns in ActivityNet Caption dataset, which leads to less than 10 objects per image.
So we apply mean pooling over the extracted object features of $t$-th segment as the $v_o^t$ feature.

The semantic feature set is to decrease the semantic gap between multimodal video representations and language.
We first select frequent nouns and verbs in the training set of ActivityNet Caption dataset, which include $C=1,106$ concepts.
Since concept annotations of each event proposal are weakly supervised where the correspondence of segments in the proposal and concepts are unknown, we formulate the concept prediction problem as a multi-instance multi-label task.
Our concept predictor is based on the above segment-level features $V_b$ and $V_o$.
For each event proposal, we evenly select $K=20$ segments and utilize the concept predictor to generate concept probabilities for the selected segments.
Max pooling is employed to obtain the proposal-level concept predictions.
We use the binary cross entropy loss to train the concept predictor.
After training, the concept predictor can generate semantic concept probabilities  $v_s^t \in \mathbb{R}^{C}$ for each segment.

The above three feature sets only represent contents in isolated segment without considering its contextual information in the whole video.
Therefore, we further propose the contextual feature set $V_c$ to enhance the segment representation with video contexts.
We train a LSTM on top of $V_b$ and $V_o$ sequences and the objective function of the LSTM is to predict concepts for each segment.
The segment-level concepts are set the same as the corresponding proposal-level concepts.
The hidden state of the LSTM is considered as context feature $v_c^t$ for the $t$-th segment.

\subsection{Event Proposal Generation}
Accuracy and coverage are both important for event proposal generation.
The proposal ranking model proposed in our previous work \cite{chen2018ruc+} carefully designs a set of features to score densely sampled proposals, which can generate proposals with high precision.
However, this approach ignores the correlation of different event proposals, which makes the top ranked proposals similar with each other and leads to low coverage. 
A recent proposed event sequence generation network (ESGN) \cite{Mun_2019_CVPR} utilizes pointer network to sequentially select event proposals step by step, which largely avoids redundancy of generated event proposals.
Though it can achieve high coverage with few proposals, the precision is inferior to that of proposal ranking model \cite{chen2018ruc+}.

Therefore, we propose to combine the two models in inference to mutually make up for their deficiencies.
We firstly train the proposal ranking model and ESGN model.
For the proposal ranking model, we rank proposals densely sampled by the sliding window approach following our previous work \cite{chen2018ruc+}.
For the ESGN model, we utilize top-80 proposals generated from the proposal ranking model as candidates and adopt algorithm in \cite{Mun_2019_CVPR} for training.
Then we utilize Algorithm~\ref{alg:event_proposal_algorithm} to generate event proposals based on the pretrained models.

\begin{algorithm}
	\caption{Inference for event proposal generation.} 
	\label{alg:event_proposal_algorithm} 
	\begin{algorithmic}[1] 
		\REQUIRE ~~\\ 
		Candidate event proposals for the video $\{p_1, \cdots, p_M\}$;\\
		Proposal ranking model $f_s(p_t)$;\\
		ESGN model $f_e(p_t|p_{<t})$; \\
		Threshold $K$;
		\ENSURE ~~\\ 
		Selected event proposals $P'$;
		\STATE initialize $P' = \{\}$,  $t=0$, $p_{<t}=\{\}$;
		\WHILE{$\argmax f_e(\cdot|p_{<t}) \neq \mathrm{EOS}$}
		\STATE $s_{p_i} = f_s(p_i) \cdot f_e(p_i|p_{<t})$;
		\STATE append $\argmax_{p_i} s_{p_i}$ into $p_{<t}$;
		\STATE append $p_i$ in top-$K$ $s_{p_i}$ into $P'$;
		\STATE $t = t+1$;
		\ENDWHILE
		\RETURN $P'$; 
	\end{algorithmic}
\end{algorithm}

\subsection{Event Caption Generation}
The event captioning models with different contexts in Section~\ref{sec:context_models} are utilized for event caption generation.
We firstly train the caption model based on groundtruth event proposals with cross entropy loss and then finetune the model with self-critical REINFORCE algorithms \cite{rennie2017self} with rewards from METEOR and CIDEr.
In order to improve the generalization of caption models on the predicted proposals, we further augment the training data with predicted event proposal whose tIoU is larger than 0.3 with groundtruth proposals.
The training caption for the predicted event proposal is the caption of the best matched groundtruth proposal.

\begin{table*}
	\centering
	\caption{Captioning performance of different captioning models on the validation set based on groundtruth event proposals.}
	\label{tab:ctx_caption}
	\begin{tabular}{c|c|ccccc|ccc|cccc} \toprule
		\multirow{2}{*}{Model} & \multirow{2}{*}{Loc} & \multicolumn{5}{c|}{Contexts} & \multicolumn{3}{c|}{Accuracy Metrics} & \multicolumn{4}{c}{Diversity Metrics}\\
		&  & $V_c$ & local & global & event & sent & BLEU & Meteor & CIDEr & SelfB & RE & SelfB2 & RE2 \\ \midrule
		\multirow{5}{*}{\begin{tabular}[c]{@{}c@{}}Intra-\\Event\end{tabular}} & $\times$ & $\times$ & $\times$ & $\times$ & $\times$ & $\times$ & 2.46 & 11.16 & 33.70 & 74.27 & 39.72 & 87.97 & 60.55 \\
		& $\times$ & \checkmark & $\times$ & $\times$ & $\times$ & $\times$ & 3.32 & 11.66 & 41.96 & 45.39 & 17.05  & 76.31 & 40.33 \\
		& $\times$ & \checkmark  & \checkmark & $\times$ & $\times$ & $\times$ & 3.78 & 11.88 & 49.42 & 37.68 & 10.93 & 66.92 & 28.73\\
		& \checkmark & \checkmark & \checkmark & $\times$ & $\times$ & $\times$ & 3.86 & \textbf{12.10} & 49.11 & 37.22 & 11.25 & 67.39 & 27.96 \\
		& \checkmark & \checkmark & \checkmark & \checkmark & $\times$ & $\times$ & 3.91 & 11.96 & 49.56 &  40.38 & 12.92 & 69.36 & 30.84\\ \midrule
		\multirow{3}{*}{\begin{tabular}[c]{@{}c@{}}Inter-\\Events\end{tabular}} & \checkmark & \checkmark & $\times$ & $\times$ & uni & \checkmark & 4.21  & 11.71 & 52.89 & 41.32 & 14.34 & 60.11 & 26.91 \\
		& \checkmark & \checkmark & $\times$ & $\times$ & uni & $\times$ & 4.35 & 11.90 & 55.89 & \textbf{36.26} & \textbf{10.59} & \textbf{54.88} & \textbf{22.34} \\
		& \checkmark & \checkmark & $\times$ & $\times$ & bi & $\times$ & \textbf{4.59} & 11.99 & \textbf{56.52} & 39.70 & 15.24 & 59.75 & 27.06  \\ \bottomrule
	\end{tabular}
\end{table*}

\subsection{Proposal and Caption Re-ranking}
In order to further improve performance, we train different captioning models and propose the following re-ranking approach to ensemble different models.

\textbf{Proposal Re-rank:}
We first re-rank all the candidate proposals and choose the top 5 as our final proposals.
We consider four factors in our proposal re-ranking, including: 
1) proposal quality from proposal generation models;
2) describability of the proposal,  which is represented by probability of generated sentence from captioning models;
3) position and 4) length.

\textbf{Caption Re-rank:}
After selecting proposals, we re-rank captions of these proposals from different caption models.
Two factors are considered in caption re-ranking, which are the number of unique words in a caption and the matching of generated words with predicted concepts $V_s$ in Section~\ref{sec:segment_feature}.
We select the best caption for each proposal.

\section{Experiments}
\label{sec:expr_results}
\subsection{Dataset}
We utilize the ActivityNet Dense Caption dataset \cite{krishna2017dense} dataset for dense video captioning, which consists of 20k videos in total with 3.7 event proposals per video on average. 
We follow the official split with 10,009 videos for training, 4,917 videos for validation and 5,044 videos for testing in the experiments except for our final testing submission.
In the final submission, we enlarge the training set with 80\% of validation set, which results in 14,009 videos for training and 917 videos for validation. 
The video in training set contains one set of event proposal segmentation while video in validation set contains two sets of proposal segmentation.

\subsection{Evaluation of Contexts for Event Captioning}

\textbf{Experimental Setting:}
In order to purely evaluate the event captioning performance, we fix event proposals as the groundtruth proposals.
We use $V_b$ as the segment-level isolated feature and $V_c$ as the segment-level contextual feature.
For the intra-event models, we set the hidden units of LSTM as 1,024, and hidden units of LSTM in inter-events models are set to be 512.
We train each model for 100 epochs and select model with best METEOR score on the validation set.

\textbf{Evaluation Metrics:} 
We evaluate the caption quality from accuracy and diversity aspects.
For the accuracy aspect, we employ the official evaluation process \cite{krishna2017dense} with tiou threshold of 0.9 since we utilize the groundtruth proposals, and evaluate on common captioning metrics including BLEU4, METEOR and CIDEr.
The higher the scores, the more accurate the captions are.
For the diversity aspect, we evaluate the Self-BLEU (SelfB) and Repetition Evaluation (RE) \cite{xiong2018move}.
The SelfB measures the similarity of each sentence against rest sentences in the video via BLEU4.
The RE computes the redundancy score of each n-gram in the video where $n=4$ in this work.
The lower the scores, the more diverse the captions of different event proposals are.
Since the validation set contains two sets of event proposals, we evaluate two types of diversity score.
The first utilizes the two sets separately and then averages on the two sets, which are denoted as SelfB and RE.
The second combines the two sets of event proposals for diversity measure, which are denoted as SelfB2 and RE2.
Since event proposals from two sets can cover similar events, the video-level diversity of the second type is supposed to be lower than the first type.

\begin{table*}
\centering
\caption{Performance of event proposal generation on the validation set. The ``\# p" denotes number of proposals.}
\label{tab:event_prop_results}
\begin{tabular}{c|c|cccc|cccc} \toprule
\multirow{2}{*}{Method} & \multirow{2}{*}{\# p} & \multicolumn{4}{c|}{Precision (@tIoU)} & \multicolumn{4}{c}{Recall (@tIoU)} \\ 
 &  & @0.3 & @0.5 & @0.7 & @0.9 & @0.3 & @0.5 & @0.7 & @0.9 \\ \midrule
PRank \cite{chen2018ruc+} & 3.08 & 99.63 & 87.06 & 52.46 & 21.51 & 60.95 & 42.08 & 26.06 & 11.66 \\
SCNG \cite{Mun_2019_CVPR} & 2.91 & 99.02 & 87.75 & 52.24 & 14.58 & 76.18 & 56.83 & 29.45 & 7.49 \\
Fusion & 2.72 & 99.28 & 86.24 & 53.14 & 21.87 & 72.24 & 50.88 & 28.71 & 10.71 \\ \bottomrule
\end{tabular}
\end{table*}

\textbf{Experimental Results:}
Table~\ref{tab:ctx_caption} shows captioning performance with different contexts based on groundtruth event proposals.
The first row in intra-event block reflects the traditional video captioning model without considering any context or location of the event, which achieves poor captioning performance on accuracy and diversity.
The second row employs the segment-level contextual information and outperforms the first row, which demonstrates the importance of context to improve captioning quality.
Explicitly encoding local context for target event captioning can further improves the captioning performance especially on the diversity.
We also find that being aware of the location information is beneficial.
However, the global context is not complementary with the above contexts and in particular deteriorates the diversity. It might result from using too many irrelevant contextual information. We will explore to dynamically attend to different contexts in our future work.

For the inter-events models, the first row in the block is similar to the captioning model proposed in \cite{Mun_2019_CVPR}, which utilizes both visual event context and textual sentence context, while the second row only employs the event context.
Our results suggest that the sentence context might not be as useful as event context in terms of accuracy and diversity.
What is more, using sentence context can slow down training and inference due to its sequential nature.
The event context is very promising to improve the diversity of event captions especially on SelfB2 and RE2 metrics.
The SelfB2 and RE2 are evaluated with two sets of event proposal segmentation which contain more similar proposals, so the lower diversity scores on these metrics indicate that the model can distinguish fine-grained events in different event segmentation.

Finally, we compare the performance of intra-event models and inter-events models enhanced by different contexts.
In terms of the accuracy measured by METEOR, intra-event models are slightly better or comparable with inter-events models.
However, the inter-events models can generate more diverse event captions than intra-event models.
Since the evaluation metric in the AcitivityNet Captioning challenge mainly considers accuracy with METEOR metric, we adopt the intra-event models as our event captioning model.
In the future, we will explore more on inter-events models for dense video captioning.




\subsection{Evaluation of Dense Video Captioning System}
\textbf{Evaluation Metrics:}
We utilize official evaluation metrics \cite{krishna2017dense} to evaluate captions of predicted proposals, which compute the caption performance for proposals possessing tiou 0.3, 0.5, 0.7 and 0.9 with the groundtruth.

\textbf{Experimental Results:}
Table~\ref{tab:event_prop_results} presents our performance for event proposal generation with $K=1$.
To be noted, the precision and recall are evaluated using the union of two sets of event proposal annotations in the validation set instead of selecting the best from the two sets \cite{Mun_2019_CVPR}.
The fusion of the two proposal generation models can balance the precision and recall better than each single model.
In Table~\ref{tab:cap_training_results}, we present the improvements from different training methods using the best intra-event model in Table~\ref{tab:ctx_caption}, which demonstrates the effectiveness of the reinforcement learning and data augmentation.
The re-ranking performance is presented in Table~\ref{tab:rerank_results}, which ensembles different captioning models trained with different combination of features and contexts.
The performance of our submitted model is presented in Table~\ref{tab:submission}.
More training data brings substantial improvement, and our model achieves 9.91 METEOR score on the testing set.

\begin{table}
\centering
\caption{Caption performance on the validation set with different training approaches based on the groundtruth proposals.}
\label{tab:cap_training_results}
\begin{tabular}{c|c|c|c} \toprule
 & CrossEntropy & +REINFORCE & +DataAugment \\ \midrule
Meteor & 12.10 & 13.97 & 14.29 \\ \bottomrule
\end{tabular}
\end{table}

\begin{table}
	\centering
	\caption{The evaluation performance of proposal and caption re-ranking on 917 validation videos. The ``\# p" denotes the number of proposals per video.}
	\label{tab:rerank_results}
	\begin{tabular}{c|c|c|c}
		\hline
	& \# p & \textbf{official} & \textbf{enlarged} \\
		\hline
		single best & 30 & 10.16 & 10.89 \\
		proposal re-rank & 5 & 10.96 (+0.80) & 11.59 (+0.70) \\
		caption re-rank & 5 & 11.46 (+0.50) & 12.24 (+0.65)\\
		\hline
	\end{tabular}
\end{table}

\begin{table}
	\centering
	\caption{The evaluation performance on the testing set of two submissions. The ``official" denotes using official split for training and ``enlarged" denotes enlarging training set.}
	\label{tab:submission}
	\begin{tabular}{c|ccc} \toprule
		& official & enlarged \\ \midrule
		METEOR & 9.0534& 9.9053  \\ \bottomrule
	\end{tabular}
\end{table}

\section{Conclusion}
\label{sec:conclusion}
In this work, we systematically evaluate contributions from different contextual information for dense video captioning.
Our preliminary experiments show that the segment-level context, local context and event context are the most beneficial contextual types.
The inter-events models are promising to generate more diverse event captions while the intra-event models are faster and achieve slightly better accuracy in terms of METEOR for event captioning. 
Our proposed system achieves significant improvements on the dense video captioning challenge.
In the future, we will explore to dynamically encode different contexts and improve intra-event and inter-events caption models.

\section{Acknowledgments}

This work was supported by National  Natural  Science Foundation of China under Grant No. 61772535 and National Key Research and Development Plan under Grant No. 2016YFB1001202. 

{\small
\bibliographystyle{unsrt}
\bibliography{reference}
}

\end{document}